\documentclass[10pt,twocolumn,letterpaper]{article}
\usepackage{cvpr}
\usepackage{times}
\usepackage{epsfig}
\usepackage{graphicx}
\usepackage{amsmath}
\usepackage{amssymb}
\usepackage{booktabs}
\usepackage{array}
\usepackage{color, colortbl}
\usepackage{footmisc}
\usepackage{soul}
\newcolumntype{C}{>{\centering\arraybackslash}p{4.165em}}
\newcolumntype{D}{>{\centering\arraybackslash}p{3.95em}}
\definecolor{LightGray}{rgb}{0.9,0.9,0.9}
\usepackage[pagebackref=true,breaklinks=true,letterpaper=true,colorlinks,bookmarks=false]{hyperref}

\cvprfinalcopy

\def\cvprPaperID{x}

\renewcommand{\baselinestretch}{.97}
\def\gvskippar{-0.5cm}

\begin{document}

\title{Learning from Synthetic Humans}

\author{
G\"{u}l Varol\footnotemark[1] \phantom{ }\footnotemark[2]
\qquad Javier Romero\footnotemark[5]
\qquad Xavier Martin\footnotemark[1] \phantom{ }\footnotemark[4]
\qquad Naureen Mahmood\footnotemark[3] \\
Inria  \qquad\qquad\phantom{ } Body Labs \phantom{ }\qquad\qquad Inria \quad\qquad\qquad 
\qquad MPI \qquad\qquad
\\ \\
\qquad Michael J. Black\footnotemark[3]
\qquad Ivan Laptev\footnotemark[1] \phantom{ }\footnotemark[2]
\qquad Cordelia Schmid\footnotemark[1] \phantom{ }\footnotemark[4]
\\
\qquad MPI \qquad\qquad\qquad Inria  \qquad\qquad\qquad Inria
}

\maketitle
\footnotetext[1]{Inria, France}
\footnotetext[2]{D\'{e}partement
	d'informatique de l'ENS, \'{E}cole normale sup\'{e}rieure, CNRS, 
	PSL Research University, 75005 Paris, France}
\footnotetext[3]{Max Planck Institute for Intelligent Systems, T\"{u}bingen, Germany}
\footnotetext[4]{Laboratoire Jean Kuntzmann, Grenoble, France}
\footnotetext[5]{Currently at Body Labs Inc., New York, NY. This work was performed
	while JR was at MPI-IS.}

\begin{abstract}
	
Estimating human pose, shape, and motion from images and videos are fundamental challenges with many applications. Recent advances in 2D human pose estimation use large amounts of manually-labeled training data for learning convolutional neural networks (CNNs). Such data is time consuming to acquire and difficult to extend. Moreover, manual labeling of 3D pose, depth and motion is impractical. In this work we present SURREAL (Synthetic hUmans foR REAL tasks): a new large-scale dataset with synthetically-generated but realistic images of people rendered from 3D sequences of human motion capture data. We generate more than 6 million frames together with ground truth pose, depth maps, and segmentation masks.  We show that CNNs trained on our synthetic dataset allow for accurate human depth estimation and human part segmentation in real RGB images. Our results and the new dataset open up new possibilities for advancing person analysis using cheap and large-scale synthetic data.

\end{abstract}

\vspace{-0.2cm}
\section{Introduction}
Convolutional Neural Networks provide significant gains to problems with large amounts of training data.
In the field of human analysis, recent datasets~\cite{andriluka14cvpr, sapp13multimodal} now
gather a sufficient number of annotated images to train networks for 2D human pose
estimation~\cite{Newell2016,wei2016convolutional}.
Other tasks such as accurate estimation of human motion, depth and body-part segmentation are
lagging behind as manual supervision for such problems at large scale is prohibitively expensive.

Images of people have rich variation in poses, clothing, hair styles, body shapes, occlusions,
viewpoints, motion blur and other factors.
Many of these variations, however, can be synthesized using existing 3D motion capture (MoCap)
data~\cite{cmu_mocap,h36m_pami} and modern tools for realistic rendering.
Provided sufficient realism, such an approach would be highly useful for many tasks as it can
generate rich ground truth in terms of depth, motion, body-part segmentation and occlusions.

\begin{figure}
\begin{center}
  \includegraphics[width=0.99\linewidth]{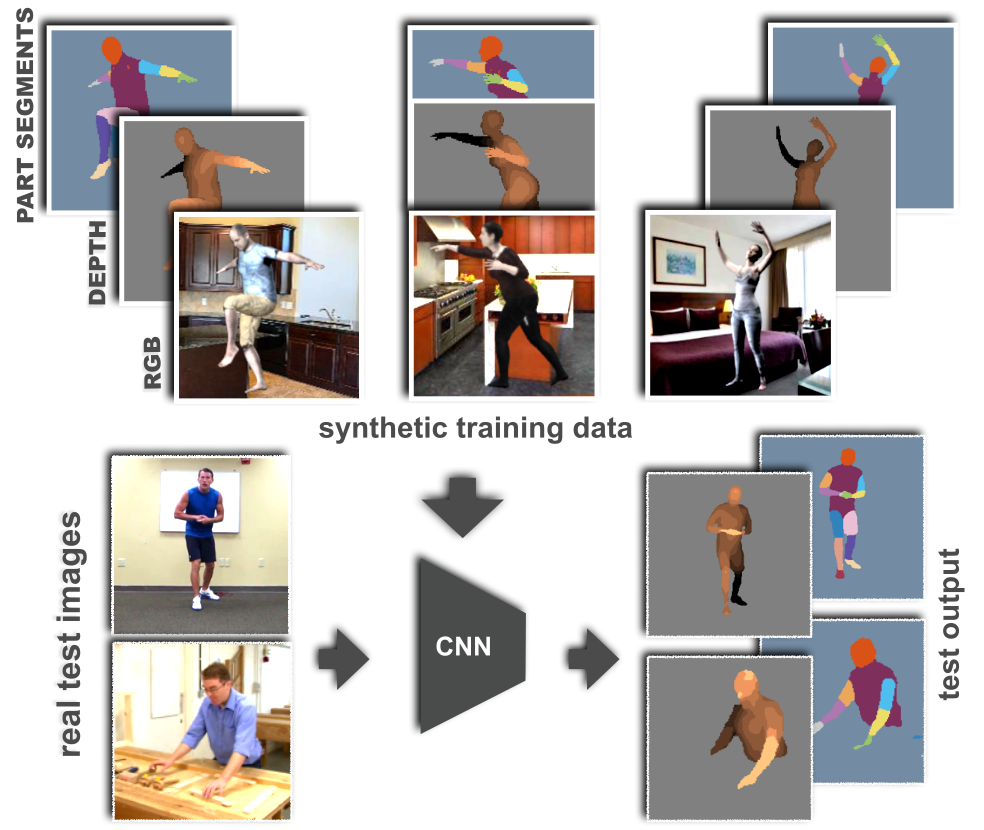}
  \mbox{}\vspace{-.3cm}\\
\end{center}
   \caption{We generate photo-realistic synthetic images and their corresponding
   ground truth for learning pixel-wise
   classification problems: human part segmentation and depth estimation. The 
   convolutional neural network trained only on synthetic data generalizes
   to real images sufficiently for both tasks. Real test images in this figure
   are taken from MPII
   Human Pose dataset~\cite{andriluka14cvpr}.}
\mbox{}\vspace{-1.3cm}\\
\label{fig:teaser}
\end{figure}

\begin{figure*}[t]
	\centering
	\includegraphics[width=0.96\textwidth,
	clip]{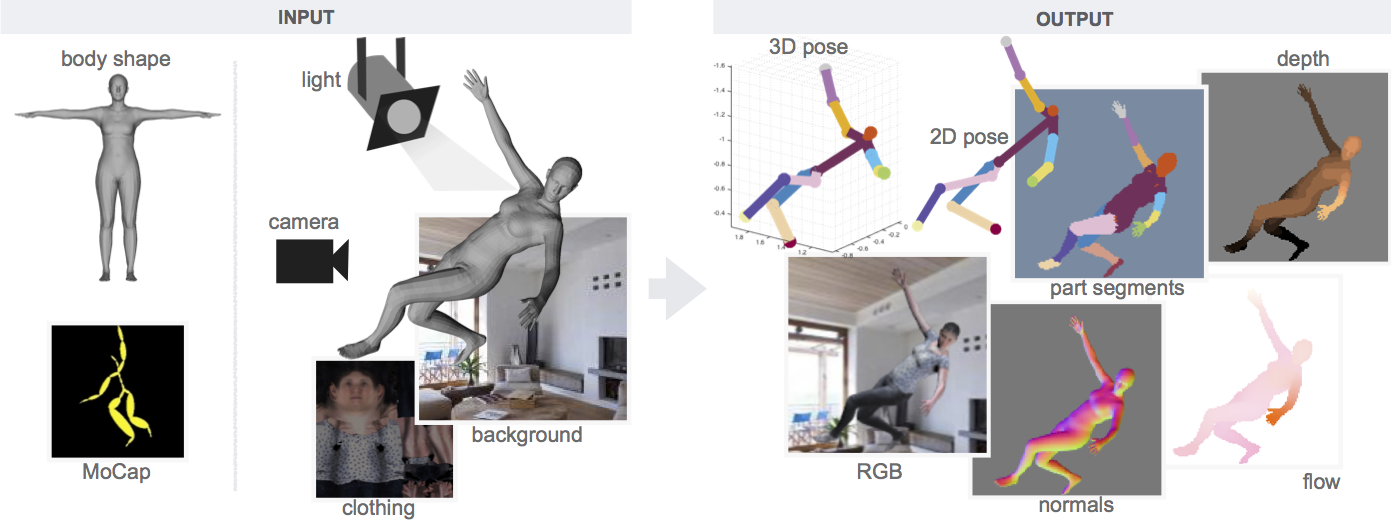}
	\caption{Our pipeline for generating synthetic data. A
		3D human body model is posed using
		motion capture data and a frame is rendered using
		a background image, a  texture map on the body,  lighting and a camera position.
		These ingredients are randomly sampled to increase the diversity of the data.
		We generate RGB images together with 2D/3D poses, surface normals, optical flow,
		depth images, and body-part segmentation maps for rendered people.}
	\mbox{}\vspace{-0.8cm}\\
	\label{fig:pipeline}
\end{figure*}
Although synthetic data has been used for many years, realism has been limited. 
In this work we present SURREAL: a new large-scale dataset
with synthetically-generated but realistic images of people.
Images are rendered from 3D sequences of MoCap data.
To ensure realism, the synthetic bodies are created
using the SMPL body model~\cite{loper_smpl},
whose parameters are fit by the MoSh~\cite{loper_mosh}
method given raw 3D MoCap marker data.
We randomly sample a large variety of viewpoints, clothing and lighting.
SURREAL contains more than 6 million frames together with ground truth
pose, depth maps, and segmentation masks.
We show that CNNs trained on synthetic data
allow for accurate human depth estimation and
human part segmentation in real RGB images, see Figure~\ref{fig:teaser}.
Here, we demonstrate that our dataset, while
being synthetic, reaches the level of realism
necessary to support training for multiple
complex tasks. This opens up opportunities for training
deep networks using graphics techniques available now.
SURREAL dataset is publicly available
together with the code to generate synthetic data and to train
models for body part segmentation and depth estimation~\cite{surrealpage}.

The rest of this paper is organized as follows. Section~\ref{sec:relatedwork}
reviews related work.
Section~\ref{sec:data} presents our approach for
generating realistic synthetic videos of people.
In Section~\ref{sec:approach} we describe our CNN architecture
for human body part segmentation and depth estimation.
Section~\ref{sec:experiments} reports experiments. We
conclude in Section~\ref{sec:conclusion}.

\section{Related work}
\label{sec:relatedwork}

Knowledge transfer from synthetic to real images has been recently studied 
with deep neural networks.
Dosovitskiy \etal~\cite{dosovitskiy_flownet} learn a CNN for optical
flow estimation using synthetically generated images of rendered 3D moving
chairs. Peng \etal~\cite{peng_synthetic} study the effect
of different visual cues such as object/background texture and
color when rendering synthetic 3D objects for object detection task. Similarly,
\cite{Su_2015_ICCV} explores rendering 3D objects to perform 
viewpoint estimation. Fanello \etal~\cite{Fanello} render
synthetic infrared images of hands and faces to predict depth and parts.
Recently, Gaidon \etal~\cite{Gaidon:Virtual:CVPR2016} have released the Virtual KITTI
dataset with synthetically generated videos of cars to study multi-object
tracking.

Several works focused on creating synthetic images of human bodies
 for learning 2D pose estimation~\cite{6248052, weichao2016, romero_flowcap},
 3D pose estimation~\cite{synthetic_cohenor, Du2016, Deep3DPose, Okada2008, 
	rogez:hal-01389486, Sminchisescu06, zhou_cvpr16}, 
pedestrian detection~\cite{MarinVGL10, 6248052, 5995574},
and action recognition~\cite{rahmani_nktm, rahmani_hpm}.
Pishchulin \etal~\cite{5995574} generate 
synthetic images with a game engine. In~\cite{6248052},
they deform 2D images with a 3D model.
More recently, Rogez and Schmid~\cite{rogez:hal-01389486}
use an image-based synthesis
engine to augment existing real images.
Ghezelghieh \etal~\cite{Deep3DPose} render synthetic images
with 10 simple body models with an emphasis on upright
people; however, the main challenge using existing MoCap
data for training is to generalize to poses that are not upright. Human3.6M dataset~\cite{h36m_pami} presents realistic rendering of people in mixed reality settings; however, the approach to create these is expensive.

A similar direction has been explored
in~\cite{rahmani_nktm, rahmani_hpm, egocap2016, shotton2011}.
In~\cite{rahmani_nktm}, action recognition is addressed 
with synthetic human trajectories from MoCap data.~\cite{rahmani_hpm, shotton2011} 
train CNNs with synthetic depth images. EgoCap~\cite{egocap2016} creates a dataset
by augmenting egocentric sequences with background.

The closest work to this paper 
is~\cite{synthetic_cohenor}, where the authors render large-scale synthetic images for 
predicting 3D pose with CNNs. Our dataset differs from~\cite{synthetic_cohenor}
by having a richer, per-pixel ground truth, thus 
allowing
to train for pixel-wise predictions and multi-task
scenarios. In addition, we argue that the realism in
our synthetic images is better (see sample videos in~\cite{surrealpage}), 
thus resulting in a smaller gap between features learned from synthetic
and real images. The method in~\cite{synthetic_cohenor}
heavily relies on real images as input
in their training with domain adaptation.
This is not the case for our synthetic training.
Moreover, we render video sequences which can be used for temporal modeling.

Our dataset presents several differences with
existing synthetic datasets.
It is the first large-scale person dataset providing depth, part segmentation and flow ground truth for synthetic RGB frames.
Other existing datasets are used either for taking RGB image as input and training only for 2D/3D pose, or for taking depth/infrared images as input and training for depth/parts segmentation.
In this paper,
we show that photo-realistic
renderings of people under large variations
in shape, texture, viewpoint and pose can help
solving pixel-wise human labeling tasks.

\begin{figure*}
	\centering
	\includegraphics[width=0.99\textwidth]{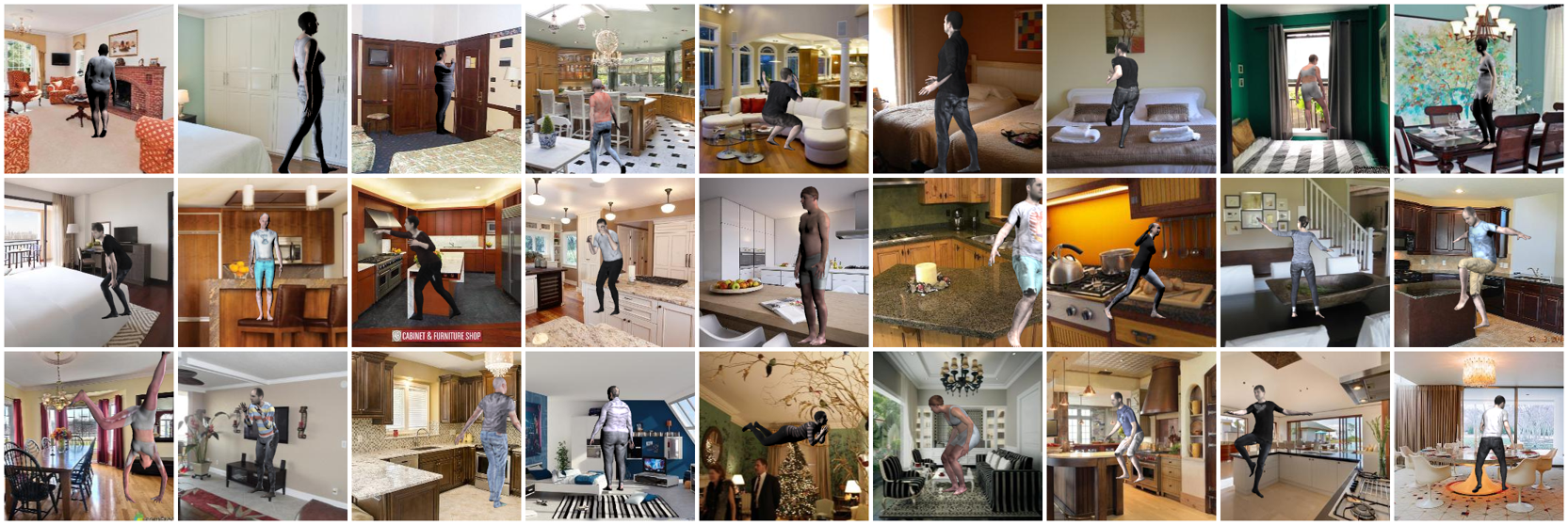}
	\caption{
		Sample frames from our SURREAL dataset with a large variety of poses, body shapes, clothings, viewpoints and backgrounds.
	}
	\mbox{}\vspace{-.65cm}\\
	\label{fig:samples}
\end{figure*}

\vspace{-0.2cm}
\section{Data generation}
\vspace{-0.1cm}
\label{sec:data}

This section presents our SURREAL (Synthetic hUmans foR REAL tasks)
dataset and describes key steps for its generation (Section~\ref{ss:synthetic_humans}).
We also describe how we obtain ground truth data for real MoCap sequences
(Section~\ref{ss:real_humans}). 

\subsection{Synthetic humans}
\label{ss:synthetic_humans}

Our pipeline for generating synthetic data is illustrated in
Figure~\ref{fig:pipeline}. 
A human body with a {\em random} 3D pose, {\em random} shape and {\em random} texture is rendered
from a {\em random} view-point for some {\em random} lighting and a {\em random} background image.
Below we define what ``random'' means in all these cases.
Since the data is synthetic, we also generate ground truth depth maps,
optical flow, surface normals, 
human part segmentations
and joint locations (both 2D and 3D).
As a result, we obtain 6.5 million frames grouped into $67,582$ continuous
image sequences. See Table~\ref{table:cmu} for more statistics, Section~\ref{subsec:synthetictest} for the description of the synthetic
train/test split, and Figure~\ref{fig:samples} for samples from the SURREAL dataset.

\vspace{\gvskippar}
\paragraph{Body model.} Synthetic bodies are created using
the SMPL body model~\cite{loper_smpl}. 
SMPL is a realistic articulated model of the body
created from thousands of high-quality 3D scans, which decomposes body
deformations into pose (kinematic deformations due to skeletal
posture) and shape (body deformations intrinsic to a particular
person that make them different from others). 
SMPL is compatible with most animation packages
like Blender~\cite{blender}. 
SMPL deformations are modeled as a combination of linear blend skinning and linear
blendshapes defined by principal components of body shape variation.
SMPL pose and shape parameters are converted
to a triangulated mesh using Blender, which then applies texture,
shading and adds a background to generate the final RGB output.

\vspace{\gvskippar}
\paragraph{Body shape.} In order to render varied, but
realistic, body shapes we make use of the CAESAR dataset~\cite{caeser}, which was
used to train SMPL.
To create a body shape, we select one of the CAESAR subjects at random
and approximate their shape with the first 10 SMPL shape principal
components.
Ten shape components explain
more than $95\%$ of the shape variance in CAESAR (at the resolution of
our mesh) and produce quite
realistic body shapes.

\vspace{\gvskippar}
\paragraph{Body pose.} To generate images of people in realistic poses, we
take motion capture data from the CMU MoCap database~\cite{cmu_mocap}.
CMU MoCap contains more than 2000 sequences of 23 high-level action categories,
resulting in more than 10 hours of recorded 3D locations of body markers.

It is often challenging to realistically and automatically retarget
MoCap skeleton data to a new model.
For this reason we do not use the skeleton data but rather use MoSh~\cite{loper_mosh} to
fit the SMPL parameters that best explain raw 3D MoCap marker
locations.
This gives both the 3D shape of the subject and the articulated pose
parameters of SMPL.
To increase the diversity, we replace the estimated 3D body shape with
a set of randomly sampled body shapes.

We render each CMU MoCap sequence three times using different random parameters.
Moreover, we divide the sequences
into clips of 100 frames with 30\%, 50\% and 70\% overlaps for
these three renderings. Every pose of the sequence is rendered with consistent
parameters (i.e.\ body shape, clothing, light, background etc.) within each clip.

\vspace{\gvskippar}
\paragraph{Human texture.} 
We use two types of real scans for the texture of body models.
First, we extract SMPL texture maps from CAESAR scans, which come with a color texture per 3D point.
These maps vary in skin color and person identities, however,
their quality is often low due to the low resolution,
uniform tight-fitting clothing, and visible markers placed on the face and the body.
Anthropometric markers are automatically removed from the texture images and inpainted.
To provide more variety, we extract a second set of textures
obtained from 3D scans of subjects with normal clothing.
These scans are registered with 4Cap as in~\cite{Dyna:SIGGRAPH:2015}.
The texture of real clothing substantially increases the realism of generated images,
even though  SMPL does not model 3D deformations of clothes.

$20\%$ of our data is rendered with the first set ($158$ CAESAR textures randomly
sampled from $4000$),
and the rest with the second set ($772$ clothed textures).
To preserve the anonymity of subjects, we replace
all faces in the texture maps by the average CAESAR face.
The skin color of this average face is corrected to fit the face skin color
of the original texture map. This corrected average face is blended
smoothly with the original map, resulting in a realistic and anonymized body texture.

\vspace{\gvskippar}
\paragraph{Light.}
The body is illuminated using Spherical Harmonics 
with $9$ coefficients~\cite{green2003spherical}.
The coefficients are randomly sampled
from a uniform distribution between $-0.7$ and $0.7$, apart from the ambient
illumination coefficient (which has a minimum value of $0.5$) and the vertical 
illumination component, which is biased to encourage the illumination from above.
Since Blender does not provide Spherical Harmonics illumination,
a spherical harmonic shader for the body material was implemented in
Open Shading Language.

\vspace{\gvskippar}
\paragraph{Camera.}
The projective camera has a resolution of $320\times 240$,
focal length of $60$mm and sensor size of $32$mm.
To generate images of the body in a wide range of positions, we 
take 100-frame MoCap sub-sequences and, in the first frame, render the
body so that the center of the viewport
points to the pelvis of the body, at a random distance (sampled
from a normal distribution with $8$ meters mean, $1$ meter deviation) with a random
yaw angle.
The remainder of the sequence then effectively produces bodies in a
range of locations relative to the static camera.

\vspace{\gvskippar}
\paragraph{Background.} We render the person on top of a static
background image.
To ensure that the backgrounds are reasonably realistic and do not include other people,
we sample from a subset of LSUN dataset~\cite{yu_lsun} that includes total of 400K
images from the categories kitchen, living room, bedroom and dining room.

\vspace{\gvskippar}
\paragraph{Ground truth.}
We perform multiple rendering passes in Blender to generate different types of
per-pixel ground truth. The {\em material} pass generates pixel-wise segmentation
of rendered body parts, given different material indices assigned to different
parts of our body model.
The {\em velocity} pass, typically used to simulate motion blur,
provides us with a render simulating optical flow.
The {\em depth} and {\em normal} passes, used for emulating effects like fog, bokeh
or for performing shading, produce per-pixel depth maps and normal maps.
The final texture rendering pass overlays the shaded, textured
body over the random background.
Together with this data we save camera and lighting
parameters as well as the 2D/3D positions of body joints.

\subsection{Generating ground truth for real human data}
\label{ss:real_humans}

Human3.6M dataset~\cite{IonescuSminchisescu11, h36m_pami}
provides ground truth for 2D and 3D human poses. Additionally, a subset of the dataset (H80K)~\cite{Ionescu_2014_CVPR} has segmentation annotation, but the definition of parts is different from the SMPL body parts used for our training.
We complement this ground truth and generate predicted SMPL
body-part segmentation and depth maps for people in Human3.6M for all frames.
Here again we use MoSh~\cite{loper_mosh} to fit the SMPL
body shape and pose to the raw MoCap marker data. 
This provides a good fit of the model to the shape and the pose of real bodies.
Given the provided camera calibration, we project models to images.
We then render the ground truth segmentation, depth, and 2D/3D joints
as above, while ensuring correspondence with real pixel values in the dataset. The depth is different from the time-of-flight (depth) data
provided by the official dataset.
These MoSh fits provide a form of approximate ``ground truth''. See Figures~\ref{fig:H36Msegmimg} and~\ref{fig:H36Mdepthimg} 
for generated examples. We use this for evaluation on the test set as well as for the baseline
where we train only on real data, and also for fine-tuning our
models pre-trained on synthetic data.
In the rest of the paper,
all frames from the synthetic training set are used for synthetic pre-training.

\vspace{-0.2cm}
\section{Approach}
\label{sec:approach}

\begin{table}
\centering
\caption{SURREAL dataset in numbers. Each MoCap sequence is rendered 3 times (with 3 different
overlap ratios). Clips are mostly 100 frames long. We obtain a total of 6,5 million frames.}
\label{table:cmu}
\resizebox{\linewidth}{!}{
\begin{tabular}{@{\hspace{.1in}}lrrrr@{}}
\toprule
		& \#subjects & \#sequences  & \#clips 	& 	\#frames 		\\\midrule
Train 	& 115		 & 1,964 		& 55,001	&	5,342,090		\\
Test	& 30		 & 703			& 12,528 	&	1,194,662		\\\midrule
Total	& 145 		 & 2,607 		& 67,582	&  {\bf 6,536,752} 	\\\bottomrule
\end{tabular}
}
\mbox{}\vspace{-0.5cm}\\
\end{table}

In this section, we present our approach for
human body part segmentation~\cite{chen_cvpr16, icra16} and 
human depth estimation~\cite{eigen_iccv15, eigen_nips14, liu_cvpr15},
which we train with synthetic and/or real data, see 
Section~\ref{sec:experiments} for the evaluation.

Our approach builds on the stacked hourglass network architecture
introduced originally for 2D pose estimation
problem~\cite{Newell2016}. This network involves several repetitions
of contraction followed by expansion layers which have skip
connections to implicitly model spatial relations from different
resolutions that allows bottom-up and top-down structured
prediction. The convolutional layers with residual connections and 8
`hourglass' modules are stacked on top of each  other, each successive
stack taking the previous stack's prediction as input. The reader is
referred to~\cite{Newell2016} for more details. A variant of this network has been used
for scene depth estimation~\cite{chen_nips16}. We choose this 
architecture because it can infer pixel-wise output by taking into account
human body structure.

Our network input is a
3-channel RGB image of size  $256\times256$ cropped and scaled to fit
a human bounding box using the ground truth. The network output
for each stack has dimensions $64\times64\times15$ in the case of segmentation
(14 classes plus the background)  and $64\times64\times20$ for depth
(19 depth classes plus the background). We use cross-entropy loss defined 
on all pixels for both segmentation and depth. The final loss of the
network is the sum over 8 stacks.  We train for 50K iterations for
synthetic pre-training using the RMSprop algorithm with mini-batches of
size  6 and a learning rate of $10^{-3}$. Our data augmentation during training
 includes random rotations, scaling and color jittering.

We formulate the problem as pixel-wise classification
task for both segmentation and depth.
When addressing segmentation,
each pixel is assigned to one of the pre-defined 14 human parts, namely  head, 
torso, upper legs, lower legs, upper arms, lower arms,
hands, feet (separately for right and left) or to 
the background class.
Regarding the depth, we align ground-truth depth maps
on the z-axis by the depth of the pelvis joint,
and then quantize depth values into 19 bins
(9 behind and 9 in front of the pelvis).
We set the quantization constant to 45mm to roughly
cover the depth extent of common human poses.
The network is trained to classify each pixel into one
of the 19 depth bins or background.
At test time, we first upsample feature maps of each
class with bilinear interpolation by a factor of 4 to
output the original resolution. Then, each pixel is assigned
to the class for which the corresponding channel has the 
maximum activation.

\section{Experiments}
\label{sec:experiments}

We test our approach on several datasets.
First, we evaluate the segmentation and depth estimation on the test set of our synthetic SURREAL dataset.
Second, we test the performance of segmentation on real images from the Freiburg Sitting People dataset~\cite{icra16}. 
Next, we evaluate segmentation and depth estimation on real videos from the Human3.6M dataset~\cite{IonescuSminchisescu11, h36m_pami} with available 3D information.
Then, we qualitatively evaluate our approach on the more challenging MPII Human Pose dataset~\cite{andriluka14cvpr}.
Finally, we experiment and discuss design choices of the SURREAL dataset.

%%%%%%%%%%%%%%% SECTION 5.1 %%%%%%%%%%%%%%% 
\vspace{-0.1cm}
\subsection{Evaluation measures}
\vspace{-0.1cm}
\label{subsec:measures}
We use intersection over union (IOU) and pixel accuracy measures for evaluating the segmentation 
approach. The final measure is the average over 14 human parts as in~\cite{icra16}. Depth estimation is 
formulated as a classification problem, but we take into account the continuity when we evaluate. We compute
root-mean-squared-error (RMSE) between the predicted quantized depth value (class) 
and the ground truth quantized depth on the human pixels.
To interpret the error in real world coordinates, we multiply it by the quantization constant (45mm).
We also report a scale and translation invariant RMSE (st-RMSE) by solving for the best 
translation and scaling in z-axis to fit the prediction to the ground truth. Since inferring depth from 
RGB is ambiguous, this is a common technique used in evaluations~\cite{eigen_nips14}.

%%%%%%%%%%%%%%% SECTION 5.2 %%%%%%%%%%%%%%% 
\vspace{-0.1cm}
\subsection{Validation on synthetic images}
\vspace{-0.1cm}
\label{subsec:synthetictest}

\paragraph{Train/test split.} 
To evaluate our methods on synthetic images, we separate $20\%$ of the synthetic
frames for the test set and train all our networks on the remaining training set.
The split is constructed such that a given CMU MoCap subject is assigned as either train or test.
Whereas some subjects have a large number of instances, some subjects have unique actions,
and some actions are very common (walk, run, jump).
Overall, 30 subjects out of 145 are assigned as test. 28 test subjects cover all common actions,
and 2 have unique actions. Remaining subjects are used for training.
Although our synthetic images have different body shape and appearance than the subject
in the originating MoCap sequence, we still found it appropriate to split by subjects.
We separate a subset of our body shapes, clothing and background images for the test set.
This ensures that our tests are unbiased with regards to appearance, yet are still representative of all actions.
Table~\ref{table:cmu} summarizes the number of frames, clips and MoCap sequences in each split.
Clips are the continuous 100-frame sequences where we have the same random body shape, background, clothing, 
camera and lighting. A new random set is picked at every clip.
Note that a few sequences have less than 100 frames.

\begin{figure}
\small
	\begin{tabular}{DDDDD}  
	\rowcolor{LightGray}  
	Input & Pred$_{segm}$ & GT$_{segm}$ & Pred$_{depth}$ & GT$_{depth}$ \end{tabular}\\
	\includegraphics[width=0.092\textwidth]{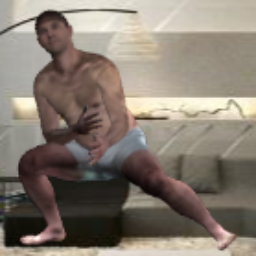}
	\includegraphics[width=0.092\textwidth]{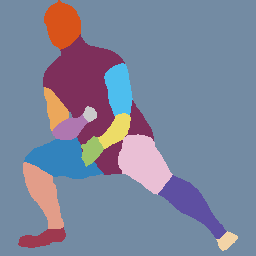} 
	\includegraphics[width=0.092\textwidth]{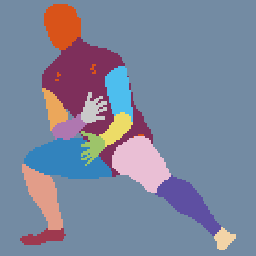} 
	\includegraphics[width=0.092\textwidth]{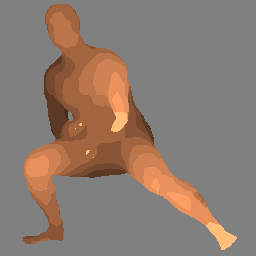} 
	\includegraphics[width=0.092\textwidth]{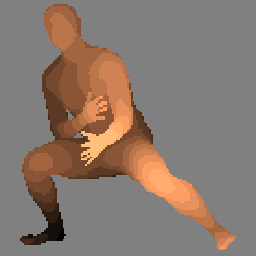} \\
	\includegraphics[width=0.092\textwidth]{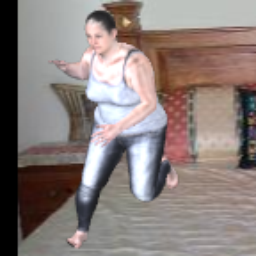}
	\includegraphics[width=0.092\textwidth]{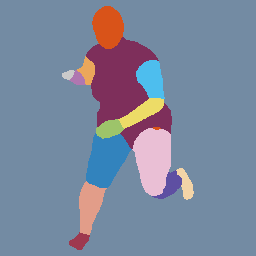} 
	\includegraphics[width=0.092\textwidth]{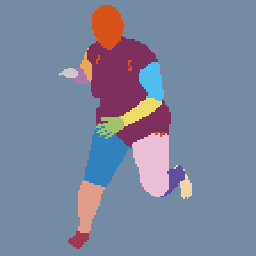} 
	\includegraphics[width=0.092\textwidth]{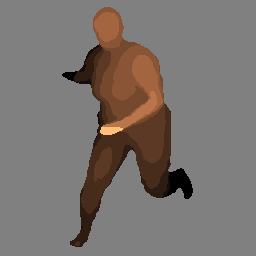} 
	\includegraphics[width=0.092\textwidth]{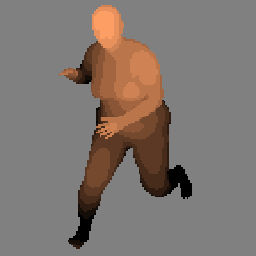}   
	\caption{Segmentation and depth predictions on synthetic test set.}
	\label{fig:cmusegmdepth}
\mbox{}\vspace{-1cm}\\
\end{figure}

\begin{figure}
\small
	\begin{tabular}{DDDDD}  
	\rowcolor{LightGray}  
	Input & Real & Synth & \hspace{-.8em}Synth+Real & GT \end{tabular}\\
	\includegraphics[width=0.092\textwidth]{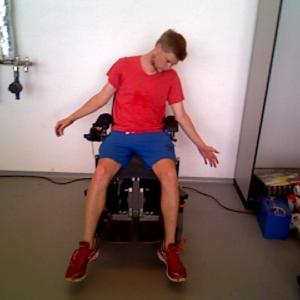}  
	\includegraphics[width=0.092\textwidth]{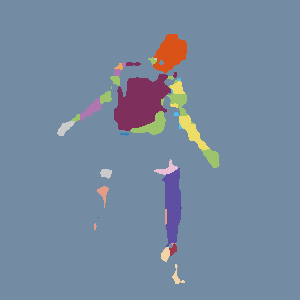} 
	\includegraphics[width=0.092\textwidth]{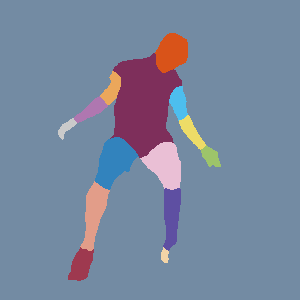}
	\includegraphics[width=0.092\textwidth]{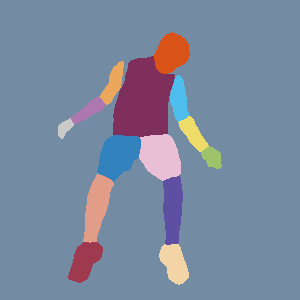}
	\includegraphics[width=0.092\textwidth]{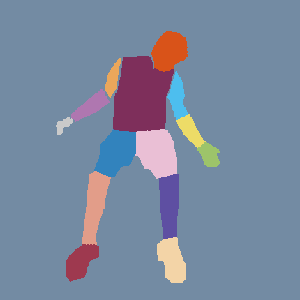}\\
	\includegraphics[width=0.092\textwidth]{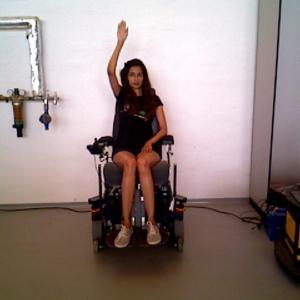} 
	\includegraphics[width=0.092\textwidth]{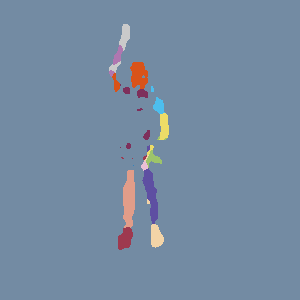}
	\includegraphics[width=0.092\textwidth]{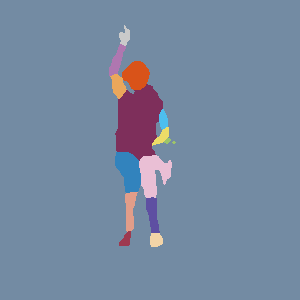}
	\includegraphics[width=0.092\textwidth]{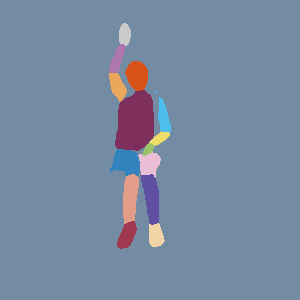}
	\includegraphics[width=0.092\textwidth]{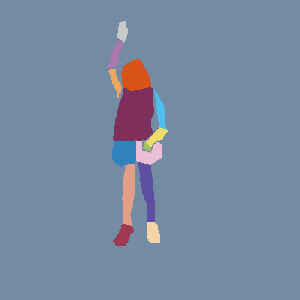} \\
	\includegraphics[width=0.092\textwidth]{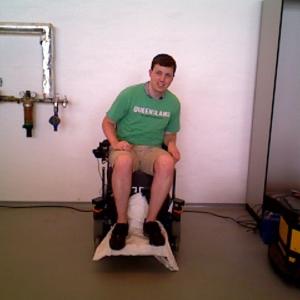} 
	\includegraphics[width=0.092\textwidth]{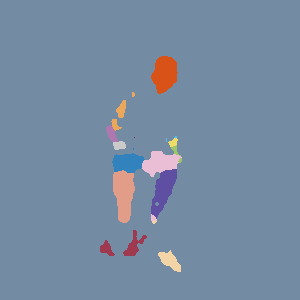}
	\includegraphics[width=0.092\textwidth]{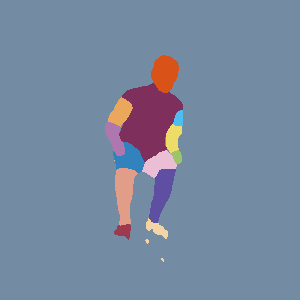}
	\includegraphics[width=0.092\textwidth]{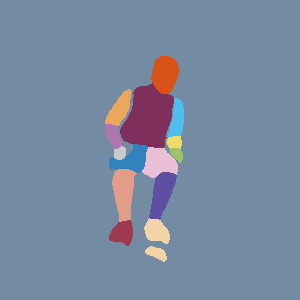}
	\includegraphics[width=0.092\textwidth]{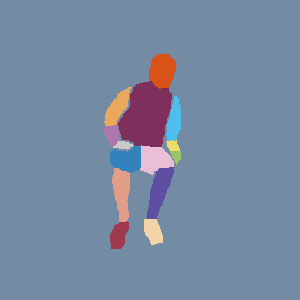} \\
	\mbox{}\vspace{-0.8cm}\\
	\caption{Part segmentation on the Freiburg Sitting People dataset, training only on 
	FSitting (Real), training only on synthetic images (Synth), fine-tuning on 2 training 
	subjects from FSitting (Synth+Real). 
	Fine-tuning helps although only 
	for 200 iterations.}
	\label{fig:FSittingsegmimg}
	\mbox{}\vspace{-1.2cm}\\
\end{figure}

\vspace{\gvskippar}
\paragraph{Results on synthetic test set.}
The evaluation is performed on the middle frame of each 100-frame clip on the aforementioned held-out synthetic 
test set, totaling in 12,528 images.
For segmentation, the IOU and pixel accuracy are 69.13\% and 
80.61\%, respectively.
Evaluation of depth estimation gives 72.9mm and
56.3mm for RMSE and st-RMSE errors, respectively.
 Figure~\ref{fig:cmusegmdepth} shows sample predictions. For
both tasks, the results are mostly accurate on synthetic test
images. However, there exist a few challenging poses (e.g. crawling),
test samples with extreme close-up views, and fine details of the hands 
that are causing errors.
In the following sections, we investigate if similar conclusions
can be made for real images.

%%%%%%%%%%%%%%% SECTION 5.3 %%%%%%%%%%%%%%% 
\vspace{-0.1cm}
\subsection{Segmentation on Freiburg Sitting People}
\vspace{-0.1cm}
Freiburg Sitting People (FSitting) dataset~\cite{icra16} is composed of 200 high resolution (300x300 
pixels) front view images of 6 subjects sitting on a wheel chair. There are 14 human part annotations 
available. See Figure~\ref{fig:FSittingsegmimg} for sample test images and corresponding ground truth 
(GT) annotation. We use the same train/test split as~\cite{icra16}, 2 subjects for training and 4 subjects 
for test.
The amount of data is limited for training deep networks. We show 
that our network pre-trained only on synthetic images is already able to segment human body parts. 
This shows that the human renderings in the synthetic dataset are representative of the real images,
such that networks trained exclusively on synthetic data can generalize quite well to real data.

\begin{table}
\centering
\caption{Parts segmentation results on 4 test subjects of Freiburg Sitting People dataset. IOU 
for head, torso and upper legs (averaged over left and right) are presented as well as 
the mean IOU and mean pixel accuracy over 14 parts. The means do not include background class. 
By adding an upsampling
layer, we get the best results reported on this dataset.}
\label{table:FSitting}
\resizebox{\linewidth}{!}{
\begin{tabular}{@{\hspace{.1in}}llllll@{}}
\toprule
						& Head 	& Torso & Legs$_{up}$ & mean 	& mean \\
Training data 			&  IOU 	& IOU 	& IOU 		  & IOU		& Acc. \\\midrule
Real+Pascal\cite{icra16}& 	-	& 	- 	&	- 	 	  & 64.10	& 81.78\\\midrule
Real					& 58.44 & 24.92 & 30.15  	  & 28.77 	& 38.02\\
Synth	 				& 73.20 & 65.55 & 39.41  	  & 40.10	& 51.88\\
Synth+Real 				& 72.88 & 80.76 & 65.41  	  & 59.58 	& 78.14\\
{\bf Synth+Real+up}	& {\bf 85.09} & {\bf 87.91} & {\bf 77.00} & {\bf 68.84} & {\bf 83.37}\\\bottomrule
\vspace{-0.8cm}
\end{tabular}
}
\end{table}

\begin{table}
\centering
\caption{Parts segmentation results on Human3.6M. The best result is obtained by fine-tuning
synthetic network with real images. Although the performance of the network trained only with
real data outperforms training only with synthetic, the predictions visually are worse because of
overfitting, see Figure~\ref{fig:H36Msegmimg}.}
\label{table:h36msegm}
\resizebox{\linewidth}{!}{
\begin{tabular}{@{\hspace{.1in}}l@{\hspace{1.1cm}}cc@{\hspace{1.1cm}}cc@{\hspace{.1in}}}
\toprule
 				& \multicolumn{2}{@{\hspace{-1.1cm}}c}{IOU}  & \multicolumn{2}{c}{Accuracy}  \\
Training data	& fg+bg  & fg     & fg+bg 	& fg \\\midrule
Real			& 49.61  & 46.32  & 58.54 	& 55.69 \\
Synth	 		& 46.35  & 42.91  & 56.51	& 53.55 \\
Synth+Real 		& 57.07  & 54.30  & 67.72 	& 65.53 \\\bottomrule
\vspace{-1.1cm}
\end{tabular}
}
\end{table}

Table~\ref{table:FSitting} summarizes segmentation results on FSitting. We carry out several 
experiments to understand the gain from synthetic pre-training. For the `Real' baseline, we train 
a network from scratch using 2 training subjects. This network overfits as there are  
few subjects to learn from and the performance is quite low.
Our `Synth' result is  
obtained using the network pre-trained on synthetic images without fine-tuning. 
We get 51.88\% pixel 
accuracy and 40.1\% IOU with this method and clearly outperform training from real images. 
Furthermore, fine-tuning (Synth+Real) with 2 training subjects helps significantly.
See Figure~\ref{fig:FSittingsegmimg} for qualitative results.
Given 
the little amount for training in FSitting, the fine-tuning converges after 200 iterations.

In~\cite{icra16}, the authors introduce a 
network that outputs a high-resolution segmentation after several layers of upconvolutions. For a fair 
comparison, we modify our network to output full resolution by adding one bilinear upsampling layer followed 
by 
nonlinearity (ReLU) and a convolutional layer with $3\times3$ filters that outputs $15\times300\times300$ 
instead of $15\times64\times64$ as explained in Section~\ref{sec:approach}. If we fine-tune this network
(Synth+Real+up) on FSitting, we improve performance and
outperform~\cite{icra16} by a large margin. Note that~\cite{icra16} trains on the same
FSitting training images, but added around 2,800 Pascal images. Hence they use significantly more manual
annotation than our method.

%%%%%%%%%%%%%%% SECTION 5.4 %%%%%%%%%%%%%%% 
\vspace{-0.1cm}
\subsection{Segmentation and depth on Human3.6M}
\vspace{-0.1cm}
\label{ss:res_human3.6}
To evaluate our approach, we need sufficient real data with ground truth annotations. 
Such data is expensive to obtain and currently not available. 
For this reason, we generate nearly perfect ground
truth for images recorded with a calibrated camera and given their  
MoCap data. Human3.6M~\cite{IonescuSminchisescu11, h36m_pami} is currently the largest dataset where such information is available. There 
are 3.6 million frames from 4 cameras. We use subjects S1, S5, S6, S7, S8 for training, S9 for 
validation and S11 for testing as in~\cite{rogez:hal-01389486,
  Yasin_Iqbal_CVPR2016}, but from all 4 cameras. Note that this is different from the official train/test split~\cite{h36m_pami}.
Each subject
performs each of the 15 actions twice. 
We use all frames from one of the two instances of each action for training, and
every 64$^{th}$ frame from all instances for testing. The frames have
resolution $1000\times1000$ pixels, we assume a $256\times 256$  
cropped human bounding box is given to reduce computational complexity. We evaluate the 
performance of both segmentation and depth,
and compare with the baseline for which we train a network on real images only.

\begin{figure*}\small
	\begin{tabular}{CCCCC@{\hspace{.5cm}}CCCCC}  
	\rowcolor{LightGray}  
	Input & Real & Synth & Synth+Real & GT & Input & Real & Synth & Synth+Real & GT \end{tabular}\\
	\includegraphics[width=0.096\textwidth]{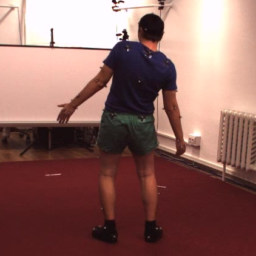}
	\includegraphics[width=0.096\textwidth]{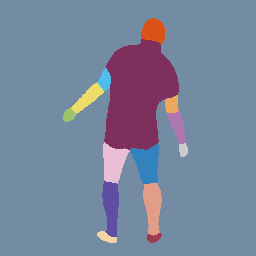}
	\includegraphics[width=0.096\textwidth]{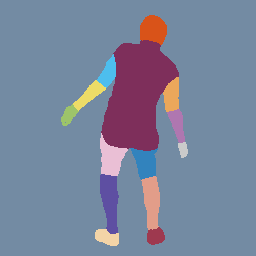}
	\includegraphics[width=0.096\textwidth]{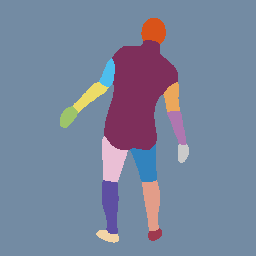}
	\includegraphics[width=0.096\textwidth]{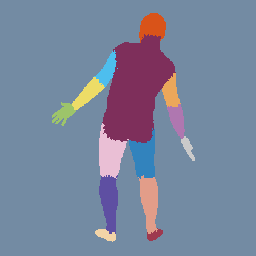}  
	\includegraphics[width=0.096\textwidth]{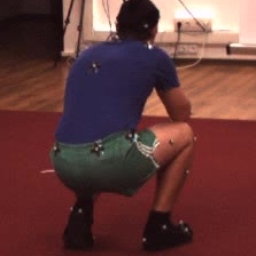} 
	\includegraphics[width=0.096\textwidth]{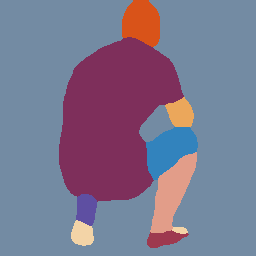} 
	\includegraphics[width=0.096\textwidth]{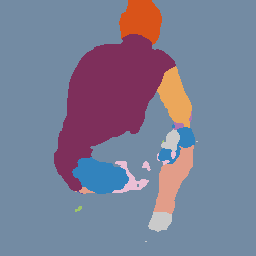} 
	\includegraphics[width=0.096\textwidth]{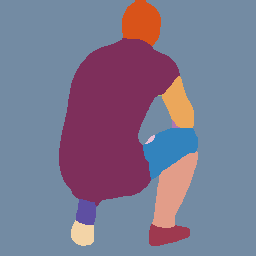} 
	\includegraphics[width=0.096\textwidth]{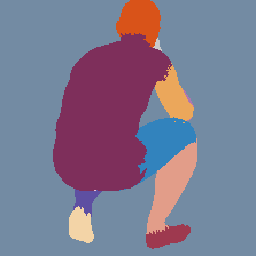}  \\ 
	\includegraphics[width=0.096\textwidth]{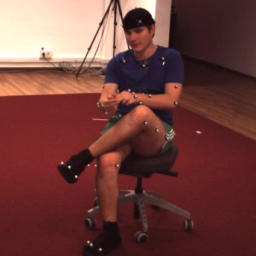}
	\includegraphics[width=0.096\textwidth]{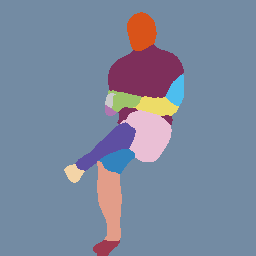}
	\includegraphics[width=0.096\textwidth]{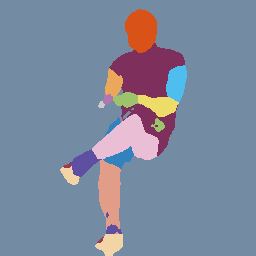}
	\includegraphics[width=0.096\textwidth]{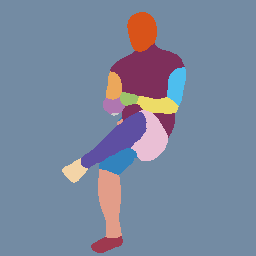}
	\includegraphics[width=0.096\textwidth]{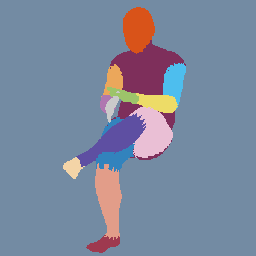}
	\includegraphics[width=0.096\textwidth]{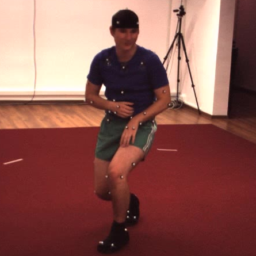}
	\includegraphics[width=0.096\textwidth]{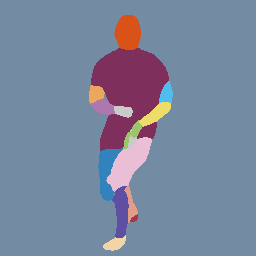}
	\includegraphics[width=0.096\textwidth]{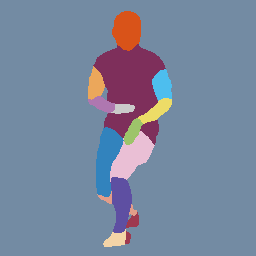}
	\includegraphics[width=0.096\textwidth]{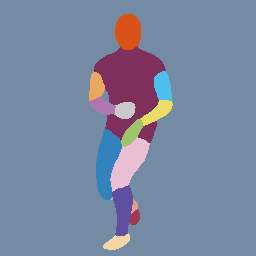}
	\includegraphics[width=0.096\textwidth]{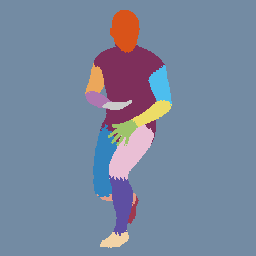} \\
	\mbox{}\vspace{-0.8cm}\\
	\caption{Parts segmentation on the Human3.6M dataset, training only on real images and 
	MoSH-generated ground-truth from Human3.6M (Real),
training only on synthetic images from SURREAL (Synth), and fine-tuning on real Human3.6M data 
(Synth+Real). The `Real' baseline clearly fails on upper arms by fitting the skin color.
The synthetic pre-trained network has seen more variety in clothing.
 Best result is achieved by the fine-tuned
network.}
\mbox{}\vspace{-1cm}\\
	\label{fig:H36Msegmimg}
\end{figure*}

\begin{figure*}
\small
	\begin{tabular}{CCCCC@{\hspace{.5cm}}CCCCC}  
	\rowcolor{LightGray}  
	Input & Real & Synth & Synth+Real & GT & Input & Real & Synth & Synth+Real & GT \end{tabular}\\
	\includegraphics[width=0.096\textwidth]{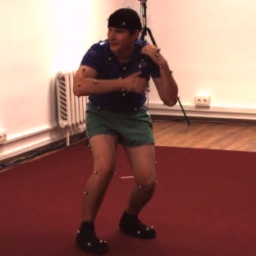}
	\includegraphics[width=0.096\textwidth]{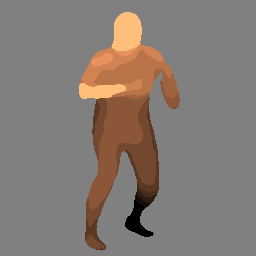}
	\includegraphics[width=0.096\textwidth]{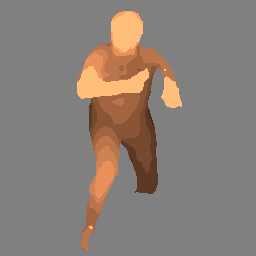}
	\includegraphics[width=0.096\textwidth]{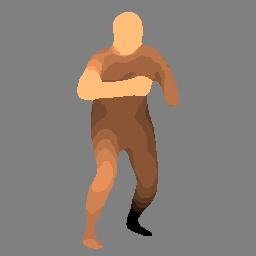}
	\includegraphics[width=0.096\textwidth]{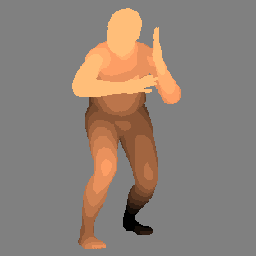}
	\includegraphics[width=0.096\textwidth]{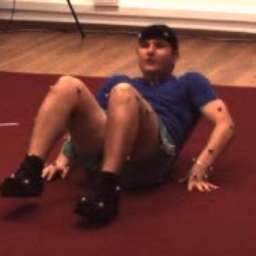}
	\includegraphics[width=0.096\textwidth]{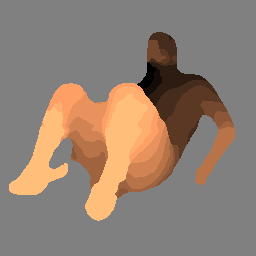}
	\includegraphics[width=0.096\textwidth]{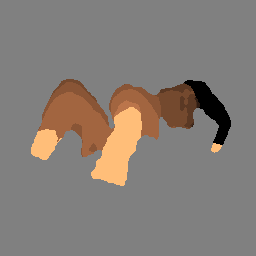}
	\includegraphics[width=0.096\textwidth]{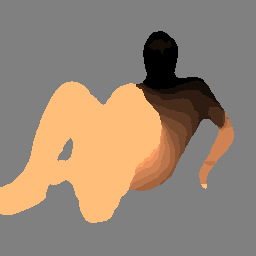}
	\includegraphics[width=0.096\textwidth]{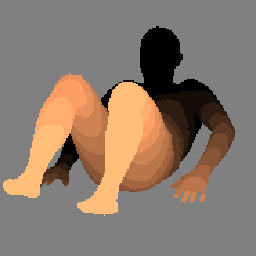} \\
	\includegraphics[width=0.096\textwidth]{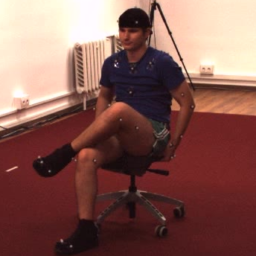}
	\includegraphics[width=0.096\textwidth]{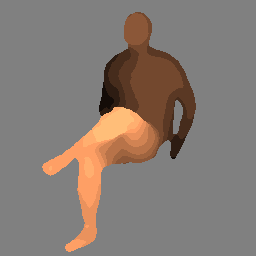}
	\includegraphics[width=0.096\textwidth]{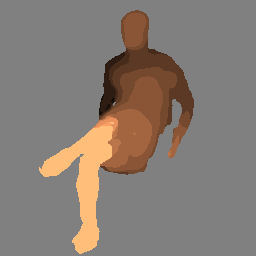}
	\includegraphics[width=0.096\textwidth]{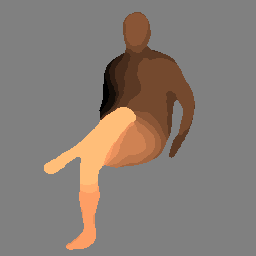}
	\includegraphics[width=0.096\textwidth]{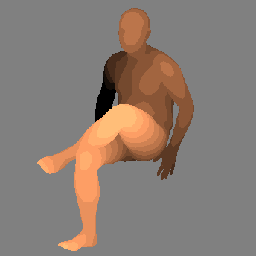} 
	\includegraphics[width=0.096\textwidth]{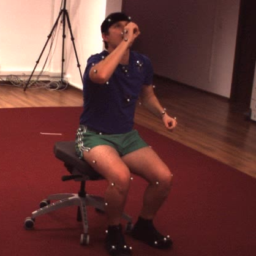}
	\includegraphics[width=0.096\textwidth]{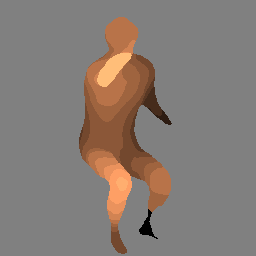}
	\includegraphics[width=0.096\textwidth]{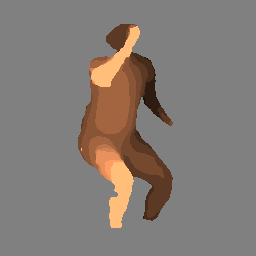}
	\includegraphics[width=0.096\textwidth]{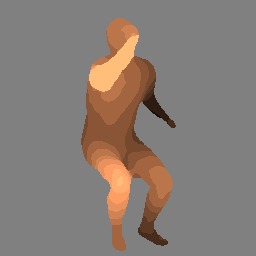}
	\includegraphics[width=0.096\textwidth]{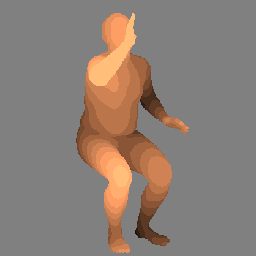}
	\mbox{}\vspace{-.8cm}\\
	\caption{Depth segmentation on the Human3.6M dataset, columns represent same training partitions as in Figure~\ref{fig:H36Msegmimg}.
The pre-trained network (Synth) fails due to scale mismatching in the training set and low contrast body parts, but fine-tuning with real data (Synth+Real) tends to recover from these problems.}
\mbox{}\vspace{-1.2cm}\\
	\label{fig:H36Mdepthimg}
\end{figure*}

\vspace{\gvskippar}
\paragraph{Segmentation.} 
Table~\ref{table:h36msegm} summarizes the parts segmentation results on Human3.6M. Note that these are not comparable to the results in~\cite{Ionescu_2014_CVPR} both because they assume the background segment is given and our ground truth segmentation data is not part of the official release (see Section~\ref{ss:real_humans}). We report both 
the mean over 14 human parts (fg) and the mean together with the background class (fg+bg).
Training on real images instead of synthetic images increases IOU by
3.4\% and pixel accuracy by 2.14\%. This is expected 
because the training distribution matches the test distribution in terms of background, camera 
position and action categories (i.e.\ poses). Furthermore, the amount of real data is sufficient to perform CNN 
training. However, since there are very few subjects available, we see that the network doesn't 
generalize to different clothing. In Figure~\ref{fig:H36Msegmimg}, the `Real' baseline has the 
border between shoulders and upper arms exactly on the T-shirt boundaries. This reveals that the 
network learns about skin color rather than actual body parts. Our pre-trained network (Synth) 
performs reasonably well, even though the pose distribution in our MoCap is quite different than that of 
Human3.6M. When we fine-tune the network with real images from Human3.6M (Synth+Real), the model
predicts very accurate segmentations and outperforms the `Real' baseline by a large margin. 
Moreover, our model is capable of distinguishing left and right most of the time on all 4 
views since it has been trained with randomly sampled views.

\vspace{\gvskippar}
\paragraph{Depth estimation.} 
Depth estimation results on Human3.6M for various poses and viewpoints are illustrated in 
Figure~\ref{fig:H36Mdepthimg}. Here, the pre-trained network fails at the very challenging poses, 
although it still captures partly correct estimates (first row). Fine-tuning
on real data compensates for these errors and refines estimations. In 
Table~\ref{table:h36mdepth}, we show RMSE error measured on foreground pixels, together with 
the scale-translation invariant version (see Section~\ref{subsec:measures}). We also report the error 
only on known 2D joints (PoseRMSE) to have an idea of how well a 3D pose estimation model would work based on 
the depth predictions. One would need to handle occluded joints to infer 3D locations of all joints, and 
this is beyond the scope of the current paper.

\begin{table}
\centering
\caption{Depth estimation results on Human3.6M (in millimeters). The depth errors 
RMSE and st-RMSE are reported on
foreground pixels. PoseRMSE error is measured only on 
given human joints.}
\label{table:h36mdepth}
\resizebox{\linewidth}{!}{
\begin{tabular}{@{\hspace{.1in}}lrrrr@{}}
\toprule
Training data	&  RMSE  & st-RMSE & PoseRMSE & st-PoseRMSE \\\midrule
Real			&  96.3  & 75.2  & 122.6 & 94.5 \\
Synth	 	& 111.6  & 98.1  & 152.5 & 131.5 \\
Synth+Real 	&  90.0  & 67.1  & 92.9 & 82.8\\\bottomrule
\vspace{-1.2cm}
\end{tabular}
}
\end{table}

\begin{figure*}%[!]
	\includegraphics[width=0.096\textwidth]{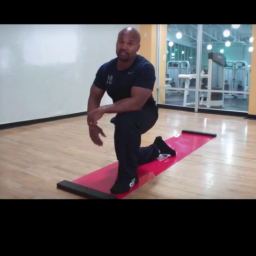}
	\includegraphics[width=0.096\textwidth]{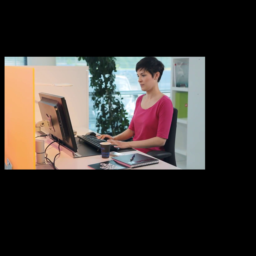}
	\includegraphics[width=0.096\textwidth]{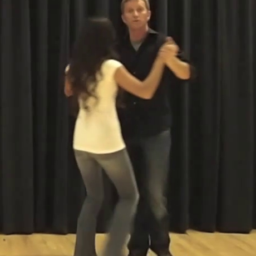}
	\includegraphics[width=0.096\textwidth]{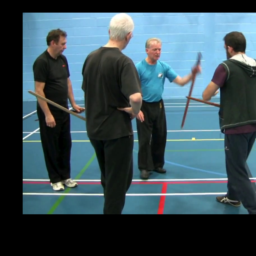}
	\includegraphics[width=0.096\textwidth]{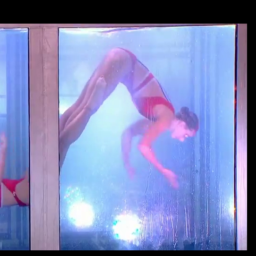}
	\includegraphics[width=0.096\textwidth]{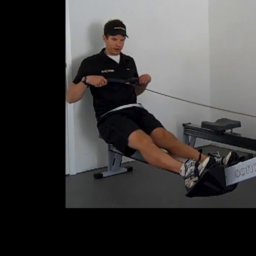}
	\includegraphics[width=0.096\textwidth]{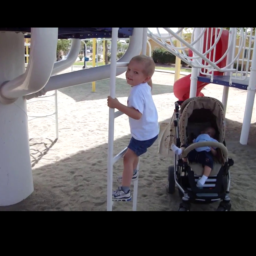}
	\includegraphics[width=0.096\textwidth]{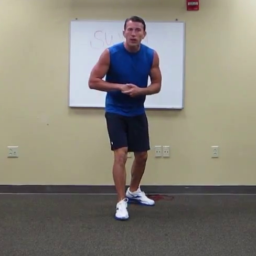}
	\includegraphics[width=0.096\textwidth]{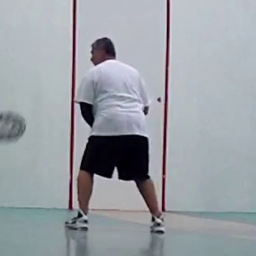}
	\includegraphics[width=0.096\textwidth]{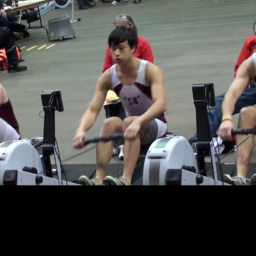}\\
	\includegraphics[width=0.096\textwidth]{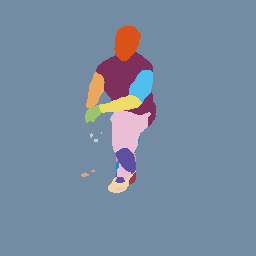}
	\includegraphics[width=0.096\textwidth]{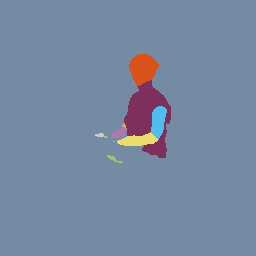}
	\includegraphics[width=0.096\textwidth]{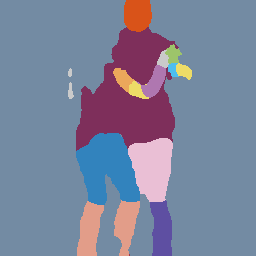}
	\includegraphics[width=0.096\textwidth]{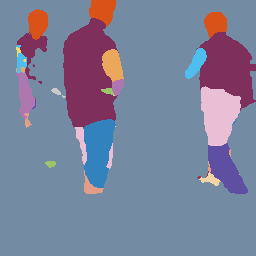}
	\includegraphics[width=0.096\textwidth]{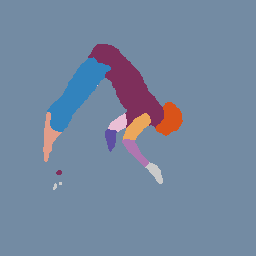}
	\includegraphics[width=0.096\textwidth]{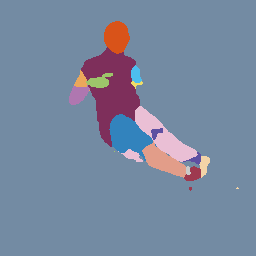}
	\includegraphics[width=0.096\textwidth]{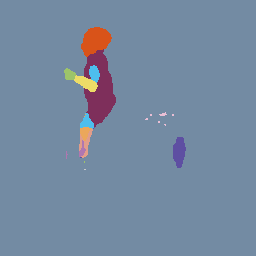}
	\includegraphics[width=0.096\textwidth]{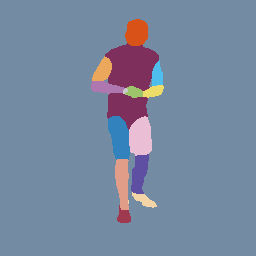}
	\includegraphics[width=0.096\textwidth]{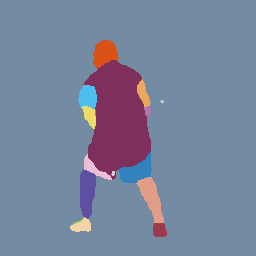}
	\includegraphics[width=0.096\textwidth]{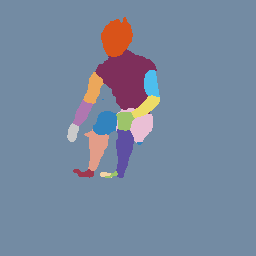}\\
	\includegraphics[width=0.096\textwidth]{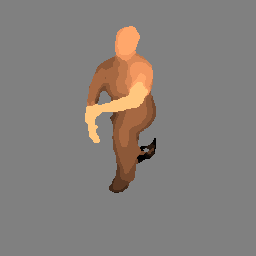}
	\includegraphics[width=0.096\textwidth]{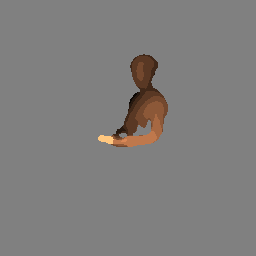}
	\includegraphics[width=0.096\textwidth]{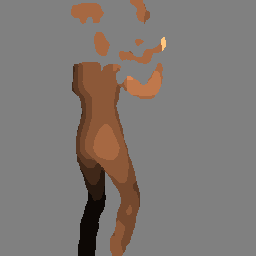}
	\includegraphics[width=0.096\textwidth]{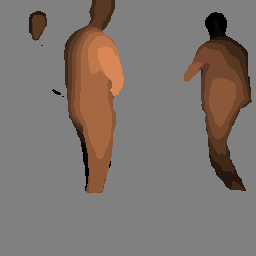}
	\includegraphics[width=0.096\textwidth]{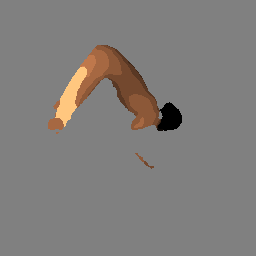}
	\includegraphics[width=0.096\textwidth]{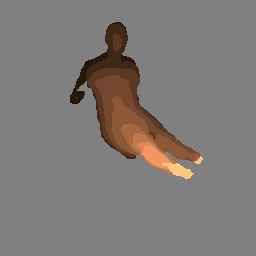}
	\includegraphics[width=0.096\textwidth]{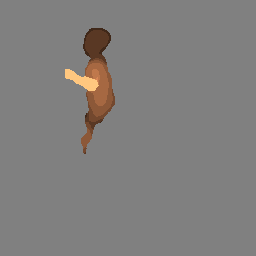}
	\includegraphics[width=0.096\textwidth]{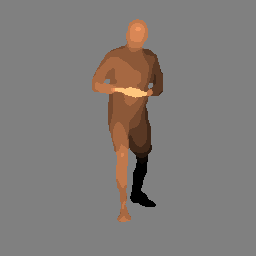}
	\includegraphics[width=0.096\textwidth]{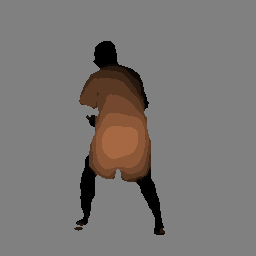}
	\includegraphics[width=0.096\textwidth]{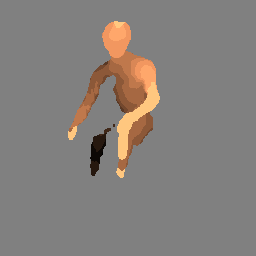} \\ \vspace{-.35cm} \\
	\includegraphics[width=0.096\textwidth]{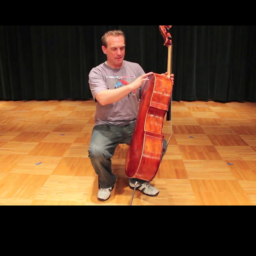}
	\includegraphics[width=0.096\textwidth]{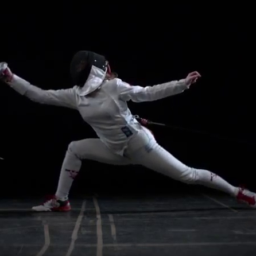}
	\includegraphics[width=0.096\textwidth]{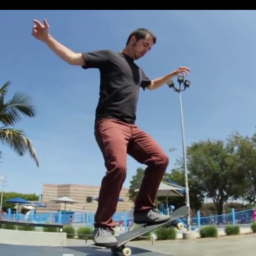}
	\includegraphics[width=0.096\textwidth]{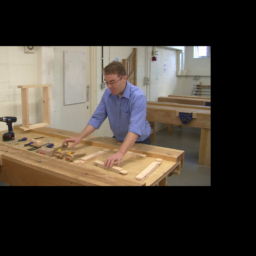}
	\includegraphics[width=0.096\textwidth]{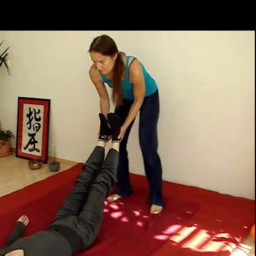}
	\includegraphics[width=0.096\textwidth]{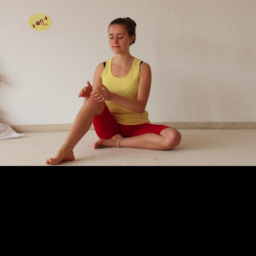}
	\includegraphics[width=0.096\textwidth]{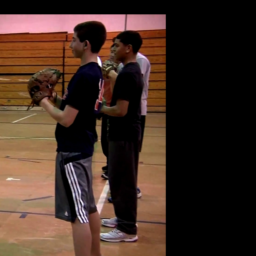}
	\includegraphics[width=0.096\textwidth]{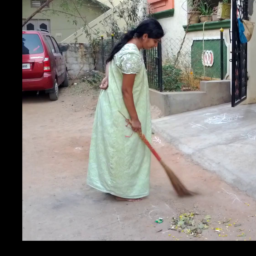}
	\includegraphics[width=0.096\textwidth]{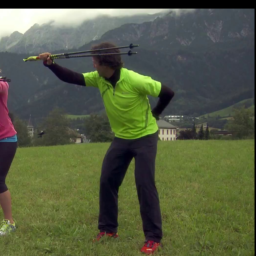}
	\includegraphics[width=0.096\textwidth]{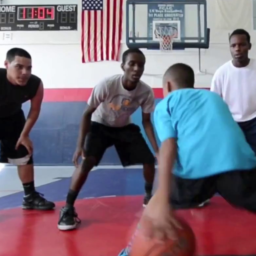}\\
	\includegraphics[width=0.096\textwidth]{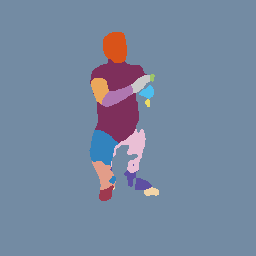}
	\includegraphics[width=0.096\textwidth]{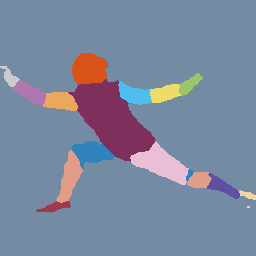}
	\includegraphics[width=0.096\textwidth]{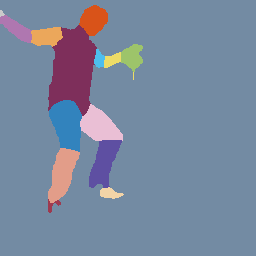}
	\includegraphics[width=0.096\textwidth]{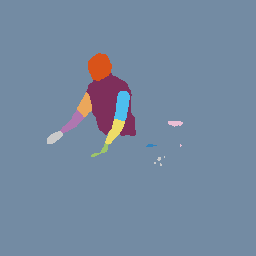}
	\includegraphics[width=0.096\textwidth]{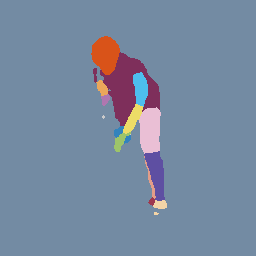}
	\includegraphics[width=0.096\textwidth]{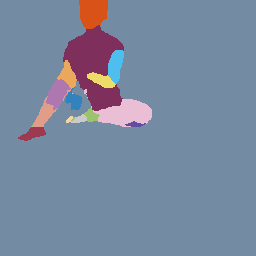}
	\includegraphics[width=0.096\textwidth]{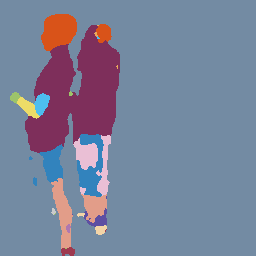}
	\includegraphics[width=0.096\textwidth]{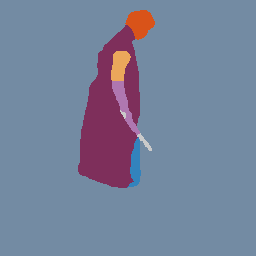}
	\includegraphics[width=0.096\textwidth]{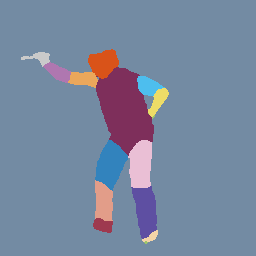}
	\includegraphics[width=0.096\textwidth]{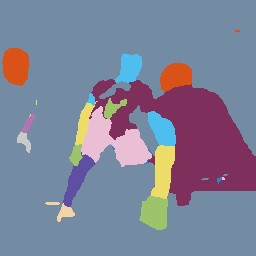}\\
	\includegraphics[width=0.096\textwidth]{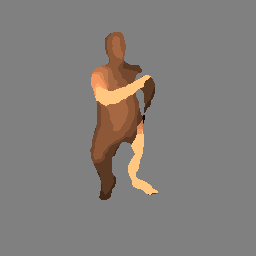}
	\includegraphics[width=0.096\textwidth]{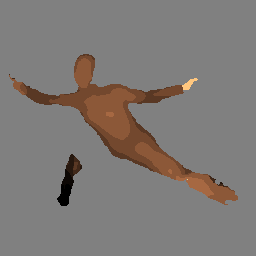}
	\includegraphics[width=0.096\textwidth]{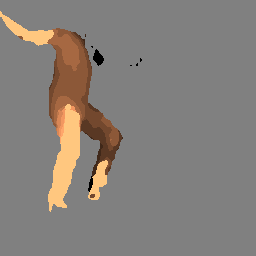}
	\includegraphics[width=0.096\textwidth]{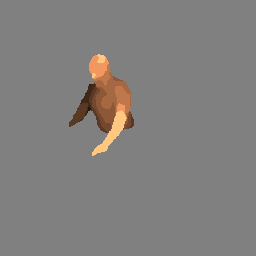}
	\includegraphics[width=0.096\textwidth]{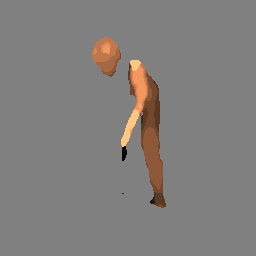}
	\includegraphics[width=0.096\textwidth]{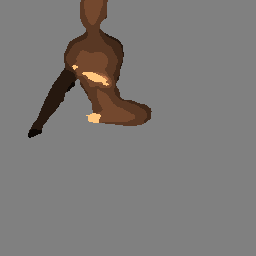}
	\includegraphics[width=0.096\textwidth]{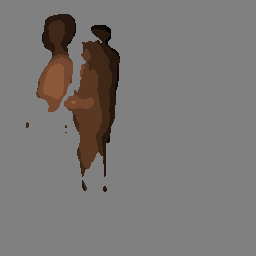}
	\includegraphics[width=0.096\textwidth]{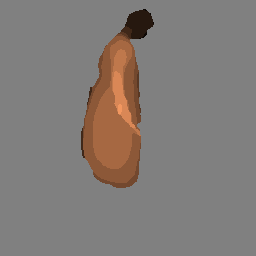}
	\includegraphics[width=0.096\textwidth]{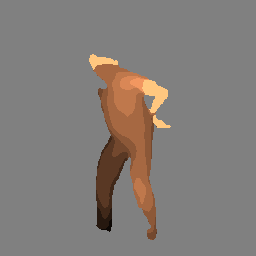}
	\includegraphics[width=0.096\textwidth]{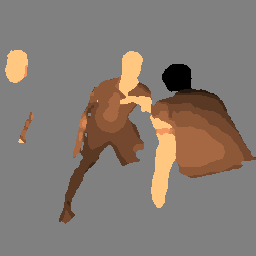}\\ 
	\mbox{}\vspace{-.85cm}\\
	\caption{Qualitative results on challenging images from MPII Human Pose dataset. 
		Multi-person, occlusion and extreme poses are difficult cases for our model. 
		Given that the model is trained only on synthetic data, it is able to generalize 
		sufficiently well on cluttered real data. It is interesting to note that although 
		we do not model cloth shape, we see in the 8th column (bottom) that 
		the whole dress is labeled as torso and depth is quite accurate. Also the lower 
		body occlusion never happens in training, but is handled well at test (2nd top, 4th 
		bottom).
	}
\mbox{}\vspace{-1.2cm}\\
	\label{fig:MPII}
\end{figure*}

%%%%%%%%%%%%%%% SECTION 5.5 %%%%%%%%%%%%%%% 
\vspace{-0.1cm}
\subsection{Qualitative results on MPII Human Pose}
\vspace{-0.1cm}
FSitting and Human3.6M are relatively simple datasets 
with limited background clutter, few subjects, single person per image, full body visible.
In this section, we test the generalization of our model
on more challenging images. MPII Human Pose~\cite{andriluka14cvpr} 
is one of the largest datasets with diverse viewpoints and clutter.
However, this dataset has no ground truth for part segmentation nor depth.
Therefore, we qualitatively show our predictions.
Figure~\ref{fig:MPII} illustrates several success and failure cases.
Our model generalizes reasonably well, except when there are multiple people close to each 
other and extreme viewpoints, which have not appeared during training. It is
interesting to note that although lower body occlusions and cloth shapes are
not present in synthetic training, the models perform accurately in such
cases, see Figure~\ref{fig:MPII} caption.

%%%%%%%%%%%%%%% SECTION 5.6 %%%%%%%%%%%%%%% 
\vspace{-0.1cm}
\subsection{Design choices}
\vspace{-0.1cm}
We did several experiments to answer questions such as `How much data 
should we synthesize?', `Is CMU MoCap enough?', `What's the effect having clothing variation?'.
 
\begin{figure}
        \mbox{}\vspace{-0.6cm}\\
	\begin{center}
		\includegraphics[width=0.47\linewidth]{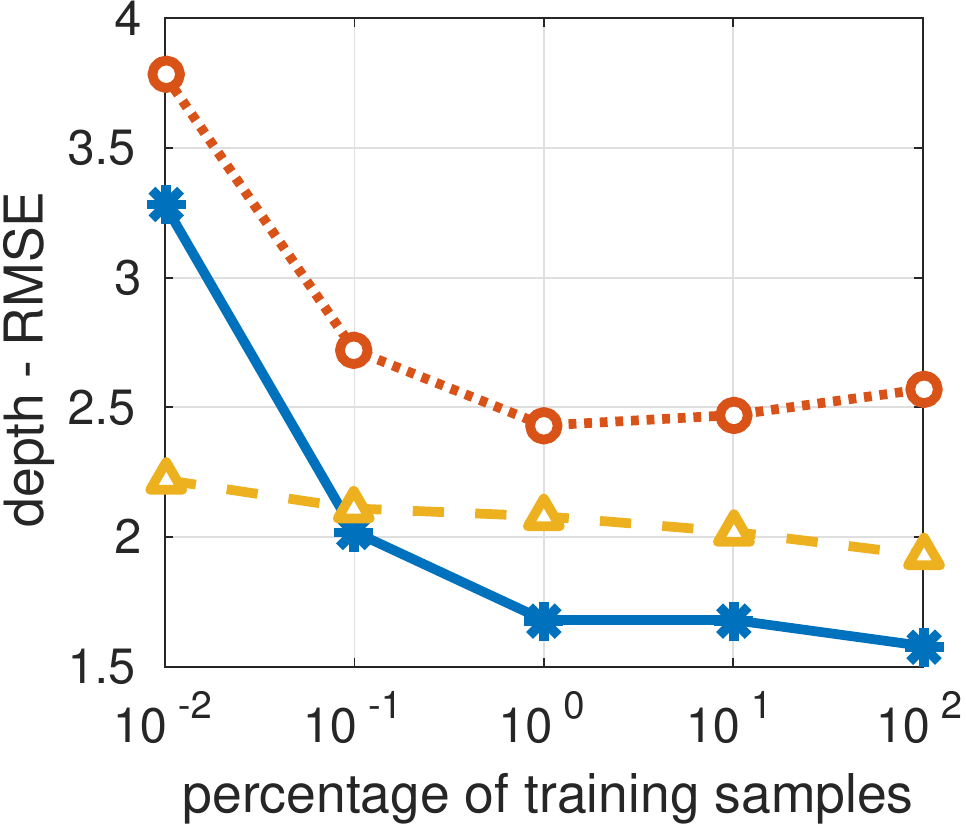} \quad
		\includegraphics[width=0.47\linewidth]{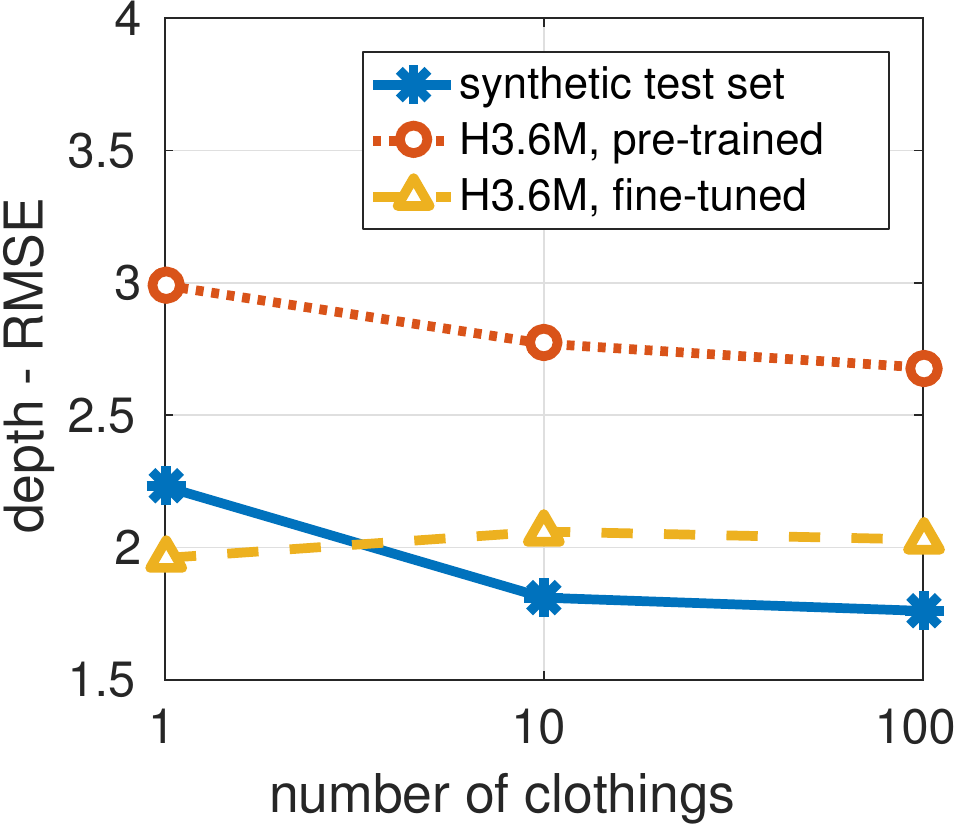}
		\includegraphics[width=0.47\linewidth]{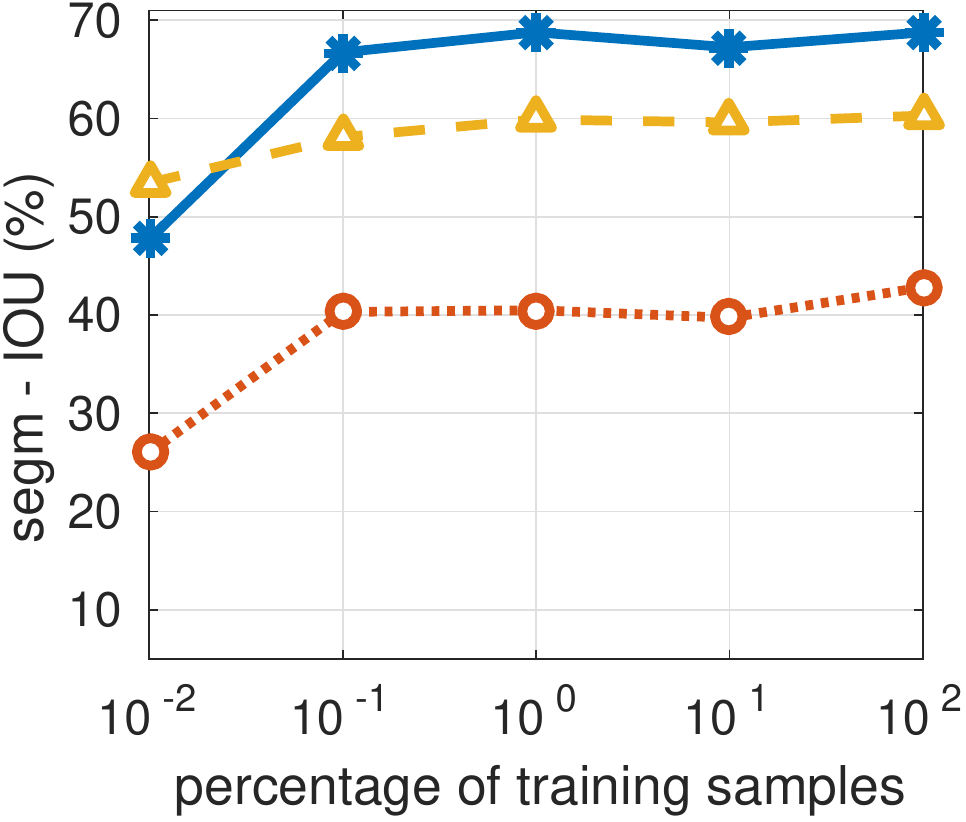} \quad
		\includegraphics[width=0.47\linewidth]{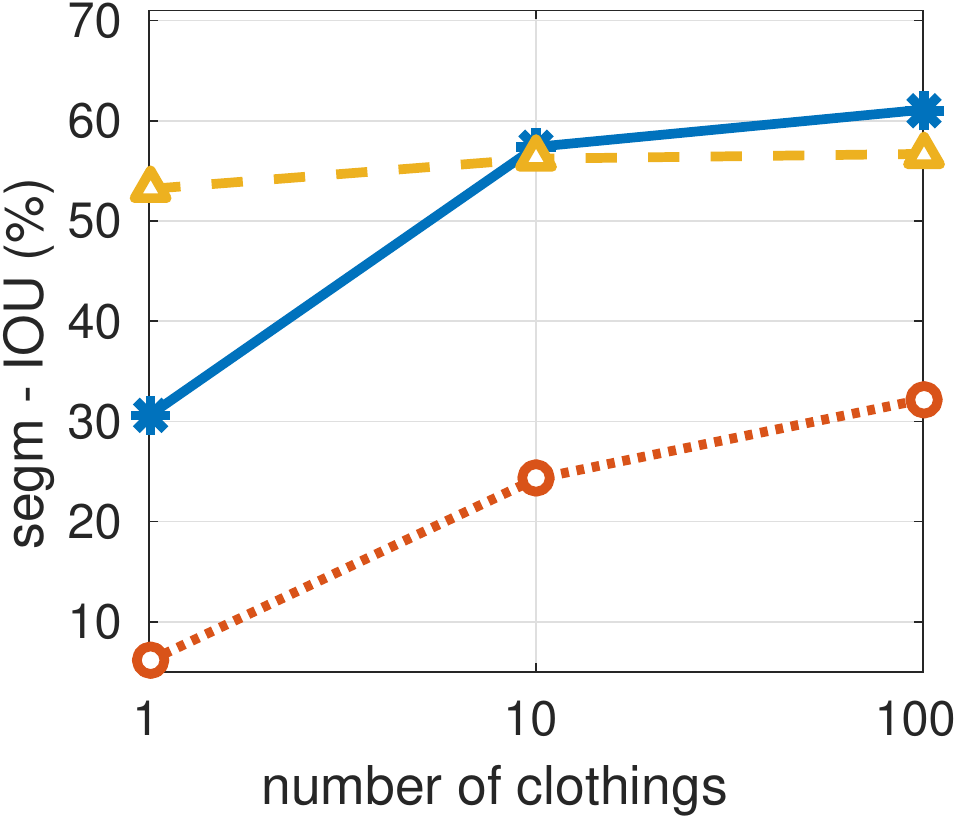}
	\end{center}
	\mbox{}\vspace{-1.2cm}\\
	\caption{\textbf{Left}: Amount of data. \textbf{Right}:
          Clothing variation. Segmentation and depth are tested on the
          synthetic and Human3.6M test sets with networks pre-trained
          on a subset of the synthetic training data. We also show
          fine-tuning on Human3.6M. The x-axis is in log-scale. }
	\label{fig:designchoices}
	\mbox{}\vspace{-1.4cm}\\
\end{figure}

\vspace{\gvskippar}
\paragraph{Amount of data.} We plot the performance as a function of training data 
size. We train with a random subset of $10^{-2}$, $10^{-1}$, $10^0$, $10^1$\% 
of the 55K training clips using all frames of the selected clips,
i.e.,~$10^{0}$\% corresponds to 550 clips with a total of 55k
frames. Figure~\ref{fig:designchoices} (left)  
shows the increase in performance for both segmentation and depth as we 
increase training data. Results are plotted on synthetic 
and Human3.6M test sets with and without fine-tuning.  
The performance gain is higher at
the beginning of all curves. There is some saturation, training with 55k
frames is sufficient, and it is more evident on Human3.6M after a
certain point. We explain this by the lack of diversity in Human3.6M
test set and the redundancy of MoCap poses. 

\vspace{\gvskippar}
\paragraph{Clothing variation.} Similarly, we study what happens when we add 
more clothing. We train with a subset of 100 clips containing only 1, 
10 or 100 different clothings (out of a total of 930),
because the dataset has maximum 100 clips for a given clothing and we want to use same number of training clips, i.e., 1 clothing with 100 clips, 10 clothings with 10 clips each and 100 clothings with 1 clip each. 
Figure~\ref{fig:designchoices} (right) shows 
the increase in performance for both tasks as we increase clothing variation. 
In the case of fine-tuning, the impact gets less prominent because training 
and test images of Human3.6M are recorded in the same room. Moreover, 
there is only one subject in our test set, ideally such experiment
should be evaluated on more diverse data.

\vspace{\gvskippar}
\paragraph{MoCap variation.} Pose distribution depends on the MoCap source.
To experiment with the effect of having similar poses in training as in test,
we rendered synthetic data using Human3.6M MoCap. 
Segmentation and depth networks pre-trained on this data (IOU: 48.11\%, 
RMSE: 2.44) outperform the ones pre-trained on CMU MoCap (42.82\%, 2.57) 
when tested on real Human3.6M.
It is important to have diverse MoCap and to match the target distribution.
Note that we exclude the Human3.6M synthetic data in Section~\ref{ss:res_human3.6} to address the more
generic case where there is no dataset specific MoCap data available.

\vspace{-0.2cm}
\section{Conclusions}
\label{sec:conclusion}
In this study, we have shown successful large-scale
training of CNNs from synthetically generated images of
 people. We have addressed two tasks, namely, human
 body part segmentation and depth estimation, for which
 large-scale manual annotation is infeasible.
 Our generated synthetic dataset 
comes with rich pixel-wise ground truth information and
can potentially be used for other tasks than considered here. 
Unlike many existing synthetic datasets, the
focus of SURREAL is on the realistic rendering of people,
which is a challenging task. In our future work, we
plan to integrate the person into the background in a more realistic
way by taking into account the lighting and the 3D scene layout.
 We also plan to augment the data with more challenging
 scenarios such as occlusions and multiple people.

\vspace{-0.2cm}

\paragraph{Acknowledgements.}
This work was supported in part by the Alexander von Humbolt Foundation, ERC
grants \mbox{ACTIVIA} and \mbox{ALLEGRO}, the MSR-Inria joint lab, and
Google and Facebook Research Awards. We acknowledge the Human3.6M dataset owners for providing the MoCap marker data.  

\renewcommand{\baselinestretch}{.96}
{\small
\bibliographystyle{ieee}
\bibliography{references}
}

\end{document}